\documentclass{article}

\usepackage{PRIMEarxiv}

\usepackage[utf8]{inputenc} % allow utf-8 input
\usepackage[T1]{fontenc}    % use 8-bit T1 fonts
\usepackage{hyperref}       % hyperlinks
\usepackage{url}            % simple URL typesetting
\usepackage{booktabs}       % professional-quality tables
\usepackage{amsfonts}       % blackboard math symbols
\usepackage{nicefrac}       % compact symbols for 1/2, etc.
\usepackage{microtype}      % microtypography
\usepackage{lipsum}
\usepackage{fancyhdr}       % header
\usepackage{graphicx}       % graphics
\graphicspath{{media/}}     % organize your images and other figures under media/ folder
\usepackage{xcolor}

\usepackage[spanish, es-tabla]{babel}
\usepackage[labelfont=bf]{caption}
\usepackage{tikz}
\usetikzlibrary{shapes.geometric, arrows.meta}
\usetikzlibrary{positioning}
\usetikzlibrary{calc}

\usepackage{apacite}

\usepackage{array}
\usepackage{multirow}
\usepackage{rotating}

\usepackage{pifont}

\title{Descripción automática de secciones delgadas de rocas: una aplicación Web}

\author{
  Stalyn Paucar, Christian Mejía-Escobar y Víctor Collaguazo\\
  \\
  Facultad de Ingeniería en Geología, Minas, Petróleos y Ambiental (FIGEMPA) \\
  Universidad Central del Ecuador \\
  Quito, Ecuador\\
  \\
  \texttt{sdpaucara1@uce.edu.ec, cimejia@uce.edu.ec, vfcollaguazo@uce.edu.ec } \\
}

\begin{document}
\maketitle

\begin{abstract}
%\lipsum[1]
La identificación y caracterización de los diversos tipos de rocas es una de las actividades fundamentales para la geología y áreas afines como la minería, el petróleo, el medio ambiente, la industria y la construcción.
Tradicionalmente, un especialista humano se encarga de analizar y explicar detalles sobre el tipo, composición, textura, forma y otras propiedades utilizando muestras de roca recogidas in-situ o preparadas en laboratorio. Los resultados se tornan subjetivos en base a la experiencia, además de consumir gran inversión de tiempo y esfuerzo.
La presente propuesta utiliza técnicas de inteligencia artificial combinando la visión por computadora y el procesamiento del lenguaje natural para generar una descripción textual y verbal a partir de una imagen de lámina delgada de roca.
Elaboramos un dataset de imágenes y sus respectivas descripciones textuales para el entrenamiento de un modelo que asocia las características relevantes de la imagen extraídas por una red neuronal convolucional EfficientNetB7 con la descripción textual generada por una red Transformer, alcanzando un valor de precisión de 0.892 y un valor BLEU de 0.71.
Este modelo puede ser un recurso útil para la investigación, el trabajo profesional y académico, por lo que ha sido desplegado a través de una aplicación Web para uso público.
\end{abstract}

%###################################################################
%#########################   INTRODUCCIÓN   ########################
%###################################################################

\keywords{Rocas \and Secciones delgadas \and Dataset de imágenes \and Inteligencia Artificial \and Deep Learning}

\section{Introducción} 
\label{sec:intro}
%Temática-importancia
Contar con información detallada sobre el tipo, composición, estructura y propiedades de las rocas es esencial no solo para la Geología y las Ciencias de la Tierra, sino también para diversas disciplinas como la exploración y extracción de recursos naturales (minerales, petróleo, gas y agua subterránea), la estabilidad del suelo, el diseño de cimientos, la planificación de proyectos de infraestructura en la ingeniería civil y la evaluación de peligros naturales (movimientos en masa, inundaciones, terremotos y tsunamis), entre otras \cite{GeologicalS}.

Una de las herramientas indispensables para los geólogos y que brinda una ventana al mundo microscópico de las rocas constituyen las denominadas \textit{secciones delgadas}. Una sección o lámina delgada es un fino corte de una muestra de roca o mineral preparado en laboratorio con un espesor estándar de $25-30 \mu m$ aproximadamente, analizado con un microscopio petrográfico \cite{raith}.
Estas secciones proporcionan información valiosa sobre la composición mineralógica, textura, tipo y otras características de las rocas. Son una herramienta complementaria para otros métodos de análisis, por ejemplo, geoquímica, fluorescencia de rayos X, activación neutrónica (INAA y RNAA), espectrometría de masas, etc. \cite{ALFEREZ2021100061}. Toda esta información ayuda a la interpretación de su formación y la comprensión de los procesos geológicos que han afectado su evolución. 

\textbf{Problemática. }
El método tradicional de identificación y caracterización de rocas a partir de secciones delgadas requiere que el observador tenga un vasto conocimiento y experiencia, lo cual involucra una fuerte carga subjetiva, un largo proceso de trabajo y fallas inherentes al mismo \cite{fan2020deep}.
En contraste, la Inteligencia Artificial (IA) se destaca en la automatización de tareas, lo que reduce significativamente el trabajo manual de manera eficiente. La capacidad de la IA para procesar grandes volúmenes de información puede ayudar a minimizar los errores humanos. Además, al basarse en algoritmos y datos, la IA tiende a minimizar la influencia de la subjetividad humana en la toma de decisiones \cite{Uribe}.

\textbf{Propuesta. }
Estos aspectos distintivos de la IA nos han motivado para llevar a cabo un sistema de descripción automática textual y verbal a partir de imágenes de secciones delgadas de roca. El objetivo de la descripción automática de imágenes es generar una oración coherente y fluida que describa con precisión el contenido de la imagen.
El auge de la IA ha despertado un interés generalizado en diversas disciplinas con el objetivo de mejorar la eficiencia en procesos de estudio e investigación.
Este enfoque ha captado la atención de la geología, que busca aprovechar estas tecnologías para optimizar la interpretación y análisis de datos geológicos, especialmente en áreas como la identificación de minerales, estructuras y texturas en muestras de rocas.

\textbf{Método. }
Recolectamos 5600 imágenes organizadas en 14 categorías de rocas, procurando elegir láminas de rocas frescas o con leve alteración, tanto para luz polarizada plana como polarizada cruzada.
Creamos las respectivas descripciones textuales considerando las especificaciones de tipo de roca y luz, textura, minerales o componentes principales, forma y hábito, relieve o colores de interferencia.
Así conformamos un extenso dataset combinado de imágenes y texto que es la materia prima para obtener un modelo de descripción automática que conecta una \textit{red neuronal convolucional} (CNN) preentrenada y una red \textit{Transformer}.
Una CNN está diseñada especialmente para reconocer patrones en datos de tipo imagen y se encarga de extraer sus características, por lo que se trata de una herramienta de vanguardia en el campo de la visión por computadora \cite{wang2020cnn}. Por otra parte, el Transformer está destinado principalmente para tareas de procesamiento del lenguaje natural. Es una arquitectura de red neuronal propuesta por primera vez en el artículo ``Attention is All You Need'' de \citeA{vaswani2017attention}. En lugar de depender de recurrencias o convoluciones, el Transformer utiliza mecanismos de atención para procesar secuencias de entrada y salida de manera paralela y eficiente, convirtiéndose en la base de muchos como BERT y GPT \cite{Sanz}.
Este componente toma las características de la imagen y las descripciones como entrada y aprende a generar las mismas.
Las descripciones generadas por el modelo son evaluadas aplicando la métrica \textit{BLEU} (Bilingual Evaluation Understudy), obteniendo un índice de similitud entre la descripción automática y la descripción auténtica igual a 0.71, el cual se cataloga como aceptable.

\textbf{Contribuciones. }
Además de nuestro dataset de imágenes y descripciones, los buenos resultados en cuanto a la métrica de similitud, nos impulsan al desarrollo de una plataforma interactiva en línea donde las personas interesadas en el tema pueden cargar sus propias imágenes de láminas delgadas de rocas y evaluar las respuestas proporcionadas por el modelo tanto de manera escrita como verbal.
La implementación considera una interfaz accesible y amigable para facilitar la participación activa de la comunidad, permitiendo la retroalimentación constante del modelo a través de la diversidad de datos aportada por los usuarios.
Este enfoque colaborativo validará la utilidad práctica del modelo en diversos contextos de la geología, ya sean profesionales o académicos. El uso de esta tecnología promete mejorar la eficiencia del trabajo de campo-laboratorio, y no solo beneficia a los profesionales e investigadores geólogos y de áreas afines, sino también a los estudiantes que dispondrían en cualquier momento de un instrumento para el aprendizaje y la comprensión de este tipo de material.

Aunque la IA puede ofrecer resultados rápidos y consistentes, la experiencia humana y la interpretación contextual aún son esenciales en la comprensión profunda y la resolución de problemas geológicos.
Surge también la pregunta si la IA podría reemplazar el uso tradicional de microscopios en entornos educativos y profesionales. 
Esta respuesta dependerá de la eficacia de la tecnología en la identificación detallada de características microscópicas de las rocas.
Nuestro trabajo pretende ser un apoyo para el especialista humano y un progreso en la tarea de lograr modelos de clasificación e interpretación de rocas con precisión altamente satisfactoria, lo cual hasta el momento aún se encuentra en desarrollo \cite{chen2023rock}.

El contenido del documento se presenta de la siguiente manera: una visión general del proyecto en la Sección \ref{sec:intro}. Los trabajos relacionados más relevantes son citados en la Sección \ref{sec:stateofart}. La Sección \ref{sec:methodology} explica la metodología empleada para el desarrollo del sistema propuesto. La Sección \ref{sec:experiments} describe la parte experimental y la evaluación del modelo obtenido. Los resultados son analizados y discutidos en la Sección \ref{sec:discussion}. La creación de la aplicación Web y su puesta en práctica se presenta en la Sección \ref{sec:appweb}. Finalmente, la Sección \ref{sec:conclusion} menciona las conclusiones de nuestro trabajo y posibles líneas de investigación y desarrollo a futuro.
%Teniendo aquello en cuenta, en esta investigación se busca determinar un modelo específico y las características clave de las imágenes de láminas delgadas de rocas que posibiliten su diferenciación efectiva unas de otras.

%La precisión de los  aún no es satisfactoria debido a la poca habilidad de estos para extraer características de este tipo de imágenes \cite{chen2023rock}.
%además del análisis petrográfico se utilizan otros métodos para obtener datos adicionales y clasificar e interpretar rocas,
%La integración de la IA en la geología representa tanto un avance prometedor como un desafío continuo que requiere evaluación y adaptación constante.

%-----------------------------------------------------------------------
%-----------------------TRABAJOS RELACIONADOS---------------------------
%-----------------------------------------------------------------------

\section{Trabajos relacionados}
\label{sec:stateofart}
Se ha discutido sobre la dependencia de métodos empíricos para el estudio de secciones delgadas y la dificultad de describirlos en un lenguaje matemático \cite{xu2022deep}. 
Además, esta tarea resulta tediosa y requiere de mucho tiempo. Por tanto, es necesario desarrollar alternativas con el objetivo de agilizar esta actividad y ahorrar recursos. En este sentido, la aplicación de IA en el contexto de la geología ha generado un interés creciente. En particular, son varios los autores que han trabajado en la clasificación automática de imágenes de secciones delgadas de rocas.
A continuación, se presentan los resultados obtenidos por otros autores.

\citeA{ren2019identifying} discuten tres problemas al trabajar con imágenes de láminas delgadas de roca: la microestructura de las rocas no tienen contornos específicos como otros objetos, se componen en su mayoría por minerales de color claro y la estructura interna es desordenada. Para su modelo usaron fotos de láminas delgadas en luz polarizada plana (PPL) provenientes de la colección geológica oficial del \textit{Natural Resources Physical Geological Data Center of China Geological Survey}. Las imágenes poseen un tamaño de 300x300 píxeles y corresponden a 8 tipos de rocas. Para cada categoría se aumentó el número de imágenes usando herramientas como rotación o recorte, por lo que cada clase cuenta con 10000 imágenes. El entrenamiento y la predicción se realizaron aplicando \textit{transfer learning} con el modelo de CNN VGG16. Para verificar la veracidad de los resultados también usaron el modelo InceptionV3. Sin embargo, el primero obtuvo mejores resultados con 95.16\% de precisión.

\citeA{zhang2019intelligent} trabajan con imágenes de minerales como feldespato potásico, pertita, plagioclasa y cuarzo. Utiliza un total de 481 imágenes en luz polarizada cruzada. Para extraer características de estas imágenes usan la arquitectura de InceptionV3 con seis modelos de aprendizaje automático: Logistic Regression (LR), Support Vector Machine (SVM), Random Forest (RF), k-nearest neighbors (KNN), multilayer perceptron (MLP) y Gaussian Naive Bayes (GNB). SVM obtiene una precisión máxima de 90.6\% y GNB una precisión mínima de 78\%. Los autores proponen trabajar con más muestras minerales para entrenar el modelo, el cual sería integrado en uno que sirva para identificar rocas de forma automática. La sugerencia es válida pues son los minerales formadores de roca y su abundancia relativa la que debe ser usada para establecer el tipo de roca en particular.

Los autores \citeA{de2020petrographic} implementan cuatro tipos de CNN (VGG19, MobileNetV2, InceptionV3 y ResNet50) para la clasificación de 98 secciones delgadas bajo luz polarizada plana en cinco microfacies: limolita argilácea, limolita bioturbada, limolita calcárea, limolita calcárea porosa y limolita masiva con cemento calcáreo. Se tiene un total de 294 imágenes con resolución de 1292x968 píxeles; mediante procedimientos de aumento de datos (corte, rotación, inversión) se generaron 5515 imágenes con resolución de 644x644 píxeles para las fases de entrenamiento, validación y prueba. Todos los modelos lograron precisiones superiores al 90\%; no obstante, ResNet50 alcanzó una precisión del 96\% para el conjunto de prueba. Este estudio remarca la importancia en la implementación de los tipos de secciones delgadas en función del objetivo del proyecto. Es decir, para la clasificación de microfacies en función del tamaño de grano y el contenido de arcilla, es suficiente con imágenes de sección delgada bajo luz polarizada plana, mientras que para una clasificación de litofacies (por ejemplo, diferenciar una roca con abundancia de cuarzo de otra con abundancia de feldespato potásico) sería necesario usar imágenes bajo luz polarizada cruzada (XPL).

\citeA{polat2021automatic} reconocen la importancia de la identificación de rocas ígneas por medio de láminas delgadas, sin embargo, se presenta el inconveniente de ser un proceso que consume demasiado tiempo y que además requiere de un especialista con conocimientos en petrografía. Por ello, los autores recurren a la clasificación de rocas usando tecnologías computacionales. Trabajaron con seis tipos de rocas: andesita, basalto, dacita, latita, riolita y traquita. Para extraer las características de estas imágenes emplearon dos CNNs: DenseNet121 y ResNet50. Estos modelos fueron entrenados y validados con 1200 imágenes y 4 diferentes optimizadores (Adelta, ADAM, RMSprop y SGD). Los modelos fueron probados usando 10 iteraciones para cada optimizador, de los cuales el más prometedor fue DenseNet121 con el optimizador RMSprop que alcanzó una precisión promedio de 99.50\%. 

\citeA{xu2022deep} indican que diferentes CNNs tienen un distinto rendimiento en diversos conjuntos de datos. Utilizan siete tipos de CNNs (Xception, MobileNetV2, InceptionResNetV2, InceptionV3, Densenet121, ResNet101V2 y ResNet101) para clasificar un total de 14950 fotos de láminas delgadas en luz polarizada cruzada, distribuidas de forma irregular en 30 categorías de rocas. Se resalta que escoger la CNN adecuada optimiza los tiempos de procesamiento y mejora la precisión. Se obtiene el mejor resultado con Xception, con una precisión de 98.65\%. Además de este valor, discuten que para la evaluación de un modelo también se debe considerar su tamaño, pues se contempla la dificultad de usarlo en computadoras con diferentes características. También se mencionan problemas de identificación, siendo el más relevante en gabros y diabasas, aunque también se señalan otros en limolitas, lutitas negras, brechas volcánicas y conglomerados.

\citeA{menendez2022reconocimiento} aplican modelos VGG16, personalizado y recurrentes (RNN) en el reconocimiento y descripción automática de rocas sedimentarias: areniscas, calizas y lutitas. Utilizan imágenes de secciones delgadas bajo luz polarizada plana y cruzada, además de un archivo de texto con las descripciones de cada roca. La cantidad de imágenes procesadas fue de 900, obtenidas principalmente de la página Web del Servicio Geológico Británico (BGS). Para evaluar el modelo de descripción usan la métrica BLEU, obteniendo un 80\% de precisión con el modelo VGG16 y 74\% con el modelo personalizado, demostrando el beneficio de la implementación de transfer learning en el aprendizaje automático con respecto a una red neuronal personalizada.

\citeA{huang2023rock} presentan un modelo de clasificación para fotos macroscópicas de rocas sedimentarias (lodolita gris, carbón, arenisca, lodolita limosa, lodolita negra y limolita argilácea) utilizando un modelo de EfficientNetB7 modificado con un módulo de triple atención; este último con el objetivo de obtener información de atención espacial de las rocas, lo que mejoraría la precisión. Para demostrar la capacidad de generalización de este modelo compararon los resultados obtenidos con las redes AlexNet, GoogleNet, VGG16 y EfficientNetB7 con un módulo de atención \textit{squeeze}. A todos los modelos se les aplicó la misma estrategia de entrenamiento y configuración de hiperparámetros. En la fase de prueba, el modelo de EfficientNetB7 con módulo de triple atención obtuvo un 95\% de precisión, siendo el mejor resultado.

Gran parte de las investigaciones previas se han enfocado en la clasificación de imágenes y en la identificación de modelos de redes neuronales convolucionales más eficientes. Aunque esta perspectiva ha sido beneficiosa para abordar una parte importante del trabajo relacionado con imágenes de láminas delgadas de rocas, esta nueva investigación destaca por no limitarse a la tarea de clasificación. La contribución de este estudio está en la capacidad de obtener descripciones automáticas textuales y verbales de tipos específicos de rocas enmarcadas en las tres categorías geológicas principales: ígneas, metamórficas y sedimentarias. Así también, la implementación de una aplicación Web para estudiantes, profesionales y usuarios en general. 

%###################################################################
%######################   METODOLOGÍA   ############################
%###################################################################

\section{Metodología}
\label{sec:methodology}
\begin{figure}[!htb]
\centering
\includegraphics[width=\textwidth]{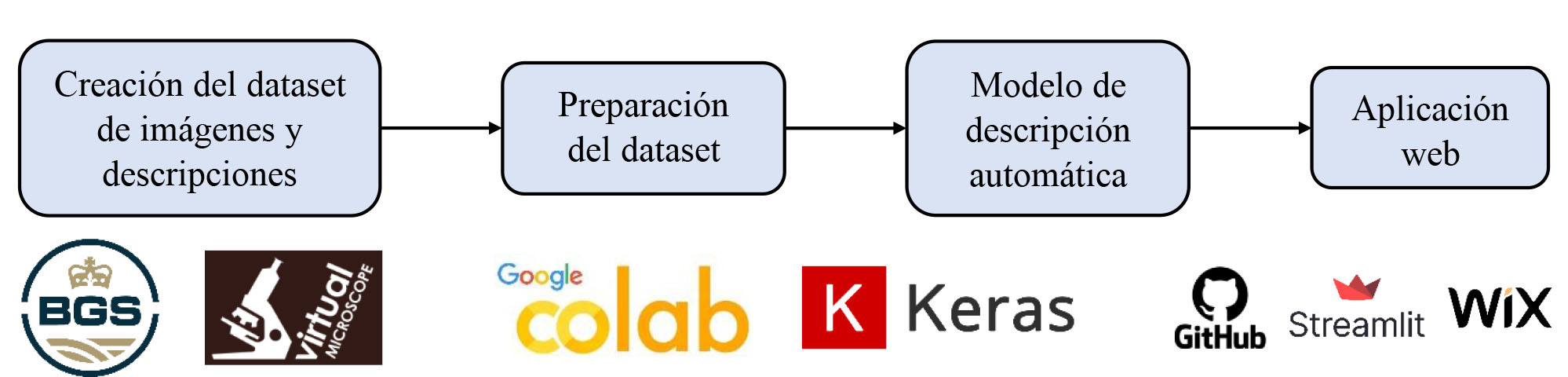}
\caption{Flujograma de las etapas principales de la metodología de trabajo, cada una con las plataformas y herramientas computacionales utilizadas para su desarrollo.}
\label{fig:methodology}
\end{figure}

La descripción de una roca proporciona una explicación de sus características esenciales a través del lenguaje y terminología geológica. Para obtenerla, los expertos realizan una inspección visual de muestras recogidas en el sitio o disponibles en laboratorio. Automatizar dicha actividad puede asistir al profesional y al estudiante, reduciendo el tiempo, el esfuerzo y la subjetividad inherente a la experiencia del ser humano. 
Nuestro trabajo tiene como objetivo la descripción textual y verbal de secciones delgadas de rocas de manera automática mediante técnicas de inteligencia artificial que involucran herramientas de vanguardia de la visión por computadora y el procesamiento de lenguaje natural.
En esta sección describimos la metodología propuesta, la cual comprende cuatro etapas principales representadas en la Figura \ref{fig:methodology}.

%Figura 1. Flujograma el modelo de descripción automática

En primer lugar, nos preocupamos de contar con la materia prima para los modelos de aprendizaje automático, es decir, el conjunto de datos (\textit{dataset}).
En este caso, recopilamos desde fuentes de acceso libre en la Web una cantidad significativa de imágenes de láminas delgadas de diversas clases de rocas y agregamos las descripciones textuales de cada una de las imágenes del dataset.
Segundo, el dataset es preparado convenientemente para el entrenamiento de un modelo que combina una red neuronal convolucional para la identificación de características visuales de la imagen y una red transformer para la generación de la descripción textual correspondiente. Tomamos como base un modelo ya definido en la página Web de Keras, el cual adaptamos y entrenamos utilizando la plataforma de Google Colab. Así obtenemos un modelo personalizado y específico que realiza la descripción textual automática de láminas delgadas de roca.
Como último paso, aprovechamos las plataformas de GitHub y Streamlit para implementar una aplicación e insertarla en una página Web diseñada con WIX, en donde el público en general podrá usar este modelo. Esta aplicación incluye la funcionalidad de convertir el texto a voz, lo cual mejora la accesibilidad de los usuarios. A continuación, una explicación detallada de las etapas mencionadas.

\subsection{Creación del dataset de imágenes}
\label{subsec:dataset}

El objetivo es recopilar un conjunto significativo de imágenes de láminas de rocas, mismas que sean representativas de su tipo correspondiente. Además, se requiere establecer un sistema estructurado de nomenclatura y etiquetado que facilite el desarrollo de modelos de aprendizaje automático. La iniciativa de generar un dataset diverso y de calidad se convierte en un valioso aporte para futuras investigaciones en ámbitos relacionados con el estudio de rocas.

% Recopilación
En cuanto a la obtención de imágenes de láminas delgadas de rocas, existen diversas fuentes de carácter público como atlas digitales o físicos, plataformas de microscopios virtuales, bases de datos institucionales, artículos científicos, tesis, entre otras. Es importante destacar que estas fuentes adolecen de falta de estructura y etiquetado estandarizado de manera que se puedan aprovechar directamente para aplicaciones de inteligencia artificial.

%Representatividad
Específicamente, se recolectaron 4263 imágenes de láminas de roca de \citeA{BGS2024} y \citeA{OpenUniversity2023}, mientras que 1337 imágenes se obtuvieron de atlas de petrografía (\citeA{mackenzie1982igneous}, \citeA{mackenzieatlascolor1996}, \citeA{mackenziesedimentaryatlas}, \citeA{mackenzie2017rocks}, \citeA{adams1998carbonate}) y otras páginas Web (\citeA{Strekeisen2020}, \citeA{Toronto2023}, \citeA{Mindat}, \citeA{Hollochersf}, \citeA{Derochette2021}). Así, hemos conformado un dataset de 5600 imágenes que permitirá al modelo aprender patrones más complejos y variados, mejorando su capacidad de generalización y adaptación a una amplia gama de muestras geológicas.

%Organización
Este conjunto de imágenes está organizado en las siguientes 14 categorías de rocas: riolita, andesita, basalto, granito, diorita, gabro, roca ultramáfica, filita, esquisto, gneis, mármol, arenisca, caliza y lutita. Cada una de estas categorías comprende 400 imágenes, 200 de las cuales se presentan en luz polarizada plana y 200 en luz polarizada cruzada. En este aspecto se dispone de un dataset equilibrado, lo que puede evitar un sesgo de aprendizaje durante el entrenamiento del modelo.

%Características
Entre las características más importantes del dataset, el formato es JPG, el cual se ha escogido ya que el tamaño del archivo se reduce considerablemente (3 a 4 veces menor) en comparación al archivo en formato PNG. El tamaño promedio de las imágenes es de 180 KB y una resolución promedio de 800x540 pixeles, lo cual puede ser redimensionado y estandarizado en tiempo de ejecución del código. Cabe señalar que no se incluye el nivel de aumento (zoom) con el cual se obtuvo la imagen, pues en la mayoría de casos este dato no está disponible, sin embargo, se ha considerado el suficiente campo de visión para mostrar las características texturales que diferencien una categoría de otra.

\begin{figure}[!htb]
\centering
\includegraphics[width=0.92\textwidth]{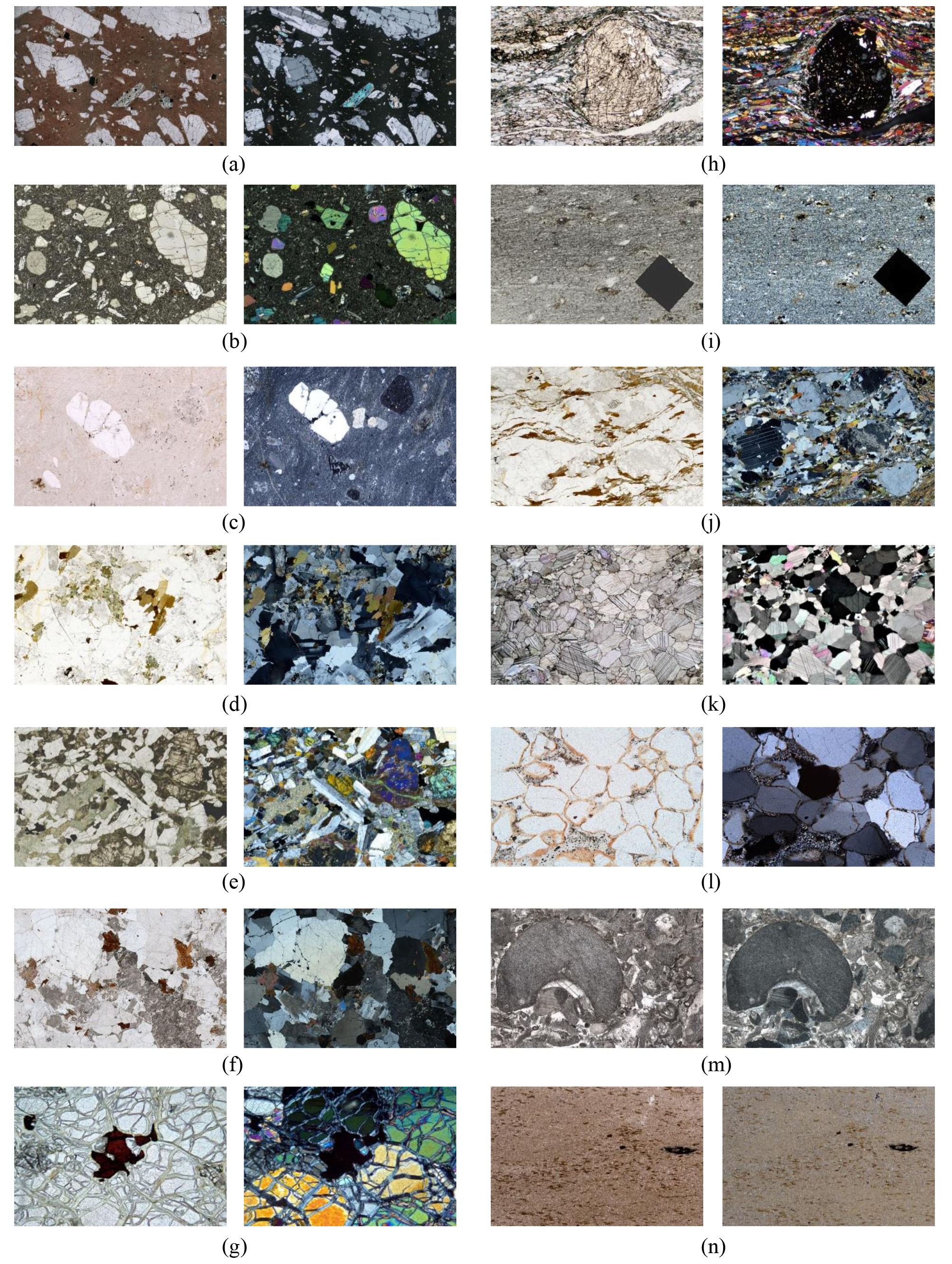}
\caption{Ejemplos de imágenes de láminas delgadas de rocas. Son pares de imágenes en luz polarizada plana y cruzada. (a) Andesita, (b) Basalto, (c) Riolita, (d) Diorita, (e) Gabro, (f) Granito, (g) Roca ultramáfica, (h) Esquisto, (i) Filita, (j) Gneis, (k) Mármol, (l) Arenisca, (m) Caliza, y (n) Lutita.}
\label{fig:muestrasrocas}
\end{figure}

% Nomenclatura y etiquetado
Los nombres asignados a cada una de las imágenes se componen de dos partes: la primera corresponde a la categoría del tipo de roca y la segunda al número de imagen de manera secuencial. La Tabla \ref{Tabla_Cod_IMG} presenta los respectivos códigos para cada categoría e imágenes, mientras que la Figura \ref{fig:muestrasrocas} muestra algunos ejemplos de las imágenes del dataset.

%Tabla con códigos de categorías e imágenes
\begin{table}[!htb]
\centering
\caption{Organización y codificación de las imágenes recolectadas.}
\renewcommand{\arraystretch}{1.4}
\begin{tabular}{ccc} 
\hline
\textbf{ Categoría } & \textbf{ Código de la categoría } & \textbf{ Código de la imagen } \\ 
\hline
\textbf{ Andesita } & 101 & 00001 - 00400 \\
\textbf{ Basalto } & 102 & 00401 - 00800 \\
\textbf{ Riolita } & 103 & 00801 - 01200 \\
\textbf{ Diorita } & 104 & 01201 - 01600 \\
\textbf{ Gabro } & 105 & 01601 - 02000 \\
\textbf{ Granito } & 106 & 02001 - 02400 \\
\textbf{ Roca ultramáfica } & 107 & 02401 - 02800 \\
\textbf{ Esquisto } & 108 & 02801 - 03200 \\
\textbf{ Filita } & 109 & 03201 - 03600 \\
\textbf{ Gneis } & 110 & 03601 - 04000 \\
\textbf{ Mármol } & 111 & 04001 - 04400 \\
\textbf{ Arenisca } & 112 & 04401 - 04800 \\
\textbf{ Caliza } & 113 & 04801 - 05200 \\
\textbf{ Lutita } & 114 & 05201 - 05600 \\
\hline
\end{tabular}
\label{Tabla_Cod_IMG}
\end{table}

%Figura 2. Ejemplos de imágenes

El dataset de imágenes propuesto ha sido cuidadosamente elaborado con el propósito de ser de utilidad tanto para este proyecto como para cualquier otro relacionado. Se encuentra disponible para descarga en la plataforma de \textit{Google Drive}. Es necesario realizar una solicitud a los autores para obtener acceso al mismo.

\subsection{Creación de las descripciones textuales}
\label{subsec:captioning}

Una vez que disponemos del conjunto organizado de imágenes de láminas de roca, procedemos a crear su descripción textual correspondiente. Debido a que utilizamos un aprendizaje automático de tipo supervisado, el entrenamiento del modelo aprende a partir de estas asociaciones entre cada imagen y su descripción textual. Por tanto, se debe invertir tiempo y esfuerzo en la creación manual de tales descripciones.

La descripción de secciones delgadas de rocas presenta una complejidad intrínseca, ya que implica la observación detallada de diversas propiedades, tanto en luz polarizada plana (PPL) como en luz polarizada cruzada (XPL). En el análisis con PPL, se examinan aspectos como el color, relieve, pleocroismo, clivaje y formas cristalinas. Por otor lado, en XPL, se exploran propiedades como extinción, zonación, figuras de interferencia, colores de interferencia, maclas, etc., que revelan características adicionales de los minerales presentes. Estas observaciones detalladas son necesarias para definir la composición mineral de la roca y, en consecuencia, definir el tipo e interpretar su evolución geológica. Además, la tarea se ve desafiada por limitaciones inherentes a los parámetros de descripción. Cada variedad rocosa posee sus propias características distintivas, lo que enfatiza la importancia de considerar estos detalles para obtener descripciones adecuadas. En ese sentido, se ha limitado los parámetros de descripción, los cuales son presentados en la Tabla \ref{Tabla1} de acuerdo con cada categoría de roca. Estos parámetros fueron definidos y seleccionados con base en el trabajo de \citeA{castro1991petrografia} sobre Petrografía básica.

%###################################################################
%Tabla de parámetros de descripción. 
%###################################################################
\begin{table}[!htb]
\small
\centering
\caption{Parámetros considerados para la elaboración de las descripciones textuales de las imágenes de láminas delgadas de rocas.}

\renewcommand{\arraystretch}{1.5}
\scalebox{0.95}{
\begin{tabular}{>{\centering\hspace{0pt}}m{0.18\linewidth}|>{\centering\hspace{0pt}}m{0.03\linewidth}>{\centering\hspace{0pt}}m{0.03\linewidth}>{\centering\hspace{0pt}}m{0.03\linewidth}|>{\centering\hspace{0pt}}m{0.03\linewidth}>{\centering\hspace{0pt}}m{0.03\linewidth}>{\centering\hspace{0pt}}m{0.03\linewidth}>{\centering\hspace{0pt}}m{0.03\linewidth}|>{\centering\hspace{0pt}}m{0.03\linewidth}>{\centering\hspace{0pt}}m{0.03\linewidth}>{\centering\hspace{0pt}}m{0.03\linewidth}>{\centering\hspace{0pt}}m{0.03\linewidth}|>{\centering\hspace{0pt}}m{0.03\linewidth}>{\centering\hspace{0pt}}m{0.03\linewidth}>{\centering\arraybackslash\hspace{0pt}}m{0.058\linewidth}} 

\hline
\textbf{Tipo de roca} & \multicolumn{3}{>{\centering\hspace{0pt}}m{0.15\linewidth}|}{\textbf{Ígneas}\par{}\textbf{extrusivas}} & \multicolumn{4}{>{\centering\hspace{0pt}}m{0.2\linewidth}|}{\textbf{Ígneas intrusivas}} & \multicolumn{4}{>{\centering\hspace{0pt}}m{0.2\linewidth}|}{\textbf{Metamórficas}} & \multicolumn{3}{>{\centering\arraybackslash\hspace{0pt}}m{0.15\linewidth}}{\textbf{Sedimentarias}} \\ \hline
\textcolor[rgb]{0.2,0.2,0.2}{\textbf{Parámetro}} & \begin{sideways}\textcolor[rgb]{0.2,0.2,0.2}{\textbf{Riolita}}\end{sideways} & \begin{sideways}\textbf{Andesita}\end{sideways} & \begin{sideways}\textbf{Basalto}\end{sideways} & \begin{sideways}\textbf{Granito}\end{sideways} & \begin{sideways}\textbf{Diorita}\end{sideways} & \begin{sideways}\textbf{Gabro}\end{sideways} & \begin{sideways}\textbf{Ultramáfica}\end{sideways} & \begin{sideways}\textbf{Filita}\end{sideways} & \begin{sideways}\textbf{Esquisto}\end{sideways} & \begin{sideways}\textbf{Gneis}\end{sideways} & \begin{sideways}\textbf{Mármol}\end{sideways} & \begin{sideways}\textbf{Arenisca}\end{sideways} & \begin{sideways}\textbf{Lutita}\end{sideways} & \begin{sideways}\textbf{Caliza}\end{sideways} \\ \hline

\textbf{Tipo de luz} & \ding{51} & \ding{51} & \ding{51} & \ding{51} & \ding{51} & \ding{51} & \ding{51} & \ding{51} & \ding{51} & \ding{51} & \ding{51} & \ding{51} & \ding{51} & \ding{51} \\

\textbf{Textura} & \ding{51} & \ding{51} & \ding{51} & \ding{51} & \ding{51} & \ding{51} & \ding{51} & \ding{51} & \ding{51} & \ding{51} & \ding{51} & \ding{51} & \ding{51} & \ding{51} \\

\textbf{Minerales-Componentes}\par{}\textbf{principales} & \ding{51} & \ding{51} & \ding{51} & \ding{51} & \ding{51} & \ding{51} & \ding{51} & \ding{51} & \ding{51} & \ding{51} & \ding{51} & \ding{51} & \ding{51} & \ding{51} \\

\textbf{Forma de los }\par{}\textbf{minerales} & \ding{51} & \ding{51} & \ding{51} & \ding{51} & \ding{51} & \ding{51} & \ding{51} & \ding{51} & \ding{51} & \ding{51} & \ding{51} & \ding{51} & - & - \\

\textbf{Hábito de los~}\par{}\textbf{minerales} & \ding{51} & \ding{51} & \ding{51} & \ding{51} & \ding{51} & \ding{51} & \ding{51} & \ding{51} & \ding{51} & \ding{51} & \ding{51} & - & - & - \\

\textbf{Sorteo} & - & - & - & - & - & - & - & - & - & - & - & \ding{51} & - & - \\

\textbf{Empaquetamiento} & - & - & - & - & - & - & - & - & - & - & - & \ding{51} & - & - \\

\textbf{Relieve} & \ding{51} & \ding{51} & \ding{51} & \ding{51} & \ding{51} & \ding{51} & \ding{51} & \ding{51} & \ding{51} & \ding{51} & \ding{51} & \ding{51} & \ding{51} & \ding{51} \\

\textbf{Colores de }\par{}\textbf{interferencia} & \ding{51} & \ding{51} & \ding{51} & \ding{51} & \ding{51} & \ding{51} & \ding{51} & \ding{51} & \ding{51} & \ding{51} & \ding{51} & \ding{51} & \ding{51} & \ding{51} \\ \hline

\end{tabular}
}
 \label{Tabla1}
\end{table}

Estos parámetros han sido organizados en cinco componentes que son concatenados para formar la descripción completa de cada imagen, tal como se presenta a continuación:

\begin{center}
\textit{Descripción: [Tipo de roca y luz] + [Textura] + [Minerales] + [Forma y hábito] + [Relieve o colores de interferencia]}    
\end{center}

En primer lugar, se especifica tanto el tipo de roca como el tipo de luz utilizada. En el segundo, se menciona la textura, mientras que el tercero se enfoca en los minerales o componentes principales. Luego, se describe la forma y el hábito de estos minerales. Finalmente, de acuerdo al tipo de luz se describe el relieve o los colores de interferencia. Con el fin de lograr una descripción más fluida y natural, se han incorporado conectores y signos de puntuación, lo que mejorará la calidad de la conversión de texto a voz. La Tabla \ref{Tabla2} muestra algunos ejemplos representativos de las descripciones.

%Tabla con ejemplos de descripciones
\begin{table}[!htb]
\centering
\caption{Ejemplos de descripciones textuales de rocas.}
\renewcommand{\arraystretch}{1.5}
\scalebox{0.9}{
\begin{tabular}{>{\hspace{0pt}}m{0.15\linewidth}>{\hspace{0pt}}m{0.858\linewidth}} 
\hline

\multicolumn{1}{>{\centering\hspace{0pt}}m{0.15\linewidth}}{\textbf{Etiqueta}} & \multicolumn{1}{>{\centering\arraybackslash\hspace{0pt}}m{0.85\linewidth}}{\textbf{Descripción}} \\ 

\hline
101\_00001.jpg\#0 & Se trata de una andesita en luz polarizada cruzada. Tiene textura holocristalina porfídica. Los minerales principales son piroxenos y plagioclasas. Los cristales son idiomorfos, con un hábito ecuante. Los colores de interferencia son de primer orden. \\
104\_01201.jpg\#0 & Se trata de una diorita en luz polarizada paralela. Tiene textura inequigranular alotriomórfica. Los minerales principales son plagioclasa, feldespato potásico, cuarzo y biotita, con un hábito tabular, laminar y ecuante. El relieve es fuerte. \\
108\_02801.jpg\#0 & Se trata de un esquisto visto con los nícoles paralelos. Tiene textura nematoblástica. Los minerales principales son glaucofana y cuarzo. Los cristales son subhedrales, con un hábito prismático. Tienen un relieve fuerte. \\
112\_04401.jpg\#0 & Se trata de una arenisca vista en luz polarizada paralela. Tiene textura madura. Los minerales principales son cuarzo. Un sorteo bueno. Contiene clastos redondeados y de esfericidad alta. El empaquetamiento es tipo tangente. Se nota un relieve fuerte. \\
\hline
\end{tabular}
}
\label{Tabla2}
\end{table}

Es importante destacar que existen parámetros que son descritos en todas las categorías de roca (tipo de luz, textura, minerales-componentes principales, relieve y colores de interferencia). Sin embargo, otros como sorteo y empaquetamiento son particulares de la categoría arenisca y dan mayor detalle acerca de la textura. En las categorías de lutita y caliza la forma y el hábito de minerales no se describen.

Las descripciones textuales son guardadas en un archivo de formato texto con codificación ASCII, extensión TXT y organizadas en dos columnas separadas por una tabulación. La primera columna corresponde a la etiqueta empleada para nombrar a las imágenes, mientras que la segunda equivale a las descripciones textuales con la estructura mencionada anteriormente. Cabe señalar que se requiere agregar los símbolos ‘\#0' después de cada nombre de la imagen junto con su extensión. Esta simbología permite insertar múltiples descripciones para cada imagen, pero en este proyecto sólo se asigna una descripción.

Aquí hemos expuesto el proceso de creación del dataset mixto compuesto de imágenes de láminas delgadas de roca y sus respectivas descripciones textuales. Este es el insumo requerido para el entrenamiento del modelo de descripción automática cuya estructura, elementos y funcionamiento se explican en la siguiente sección.

\subsection{Modelo de descripción automática}
\label{sec:architecture}

Nuestro trabajo tiene como base la implementación de \citeA{Nain2021}, un modelo de descripción de imágenes mediante la CNN EfficientNetB0 y un transformer. Utiliza el conocido conjunto de datos \textit{Flickr8K} que consta de más de 8 mil imágenes, cada una de ellas emparejada con cinco descripciones de foto diferentes. El código de programación forma parte de la página de ejemplos\footnote{\url{https://keras.io/examples/}} de \textit{Keras}, una librería de deep learning de código abierto creada por Google en lenguaje de programación Python.

\begin{figure}[!htb]
\centering
\includegraphics[width=0.99\textwidth]{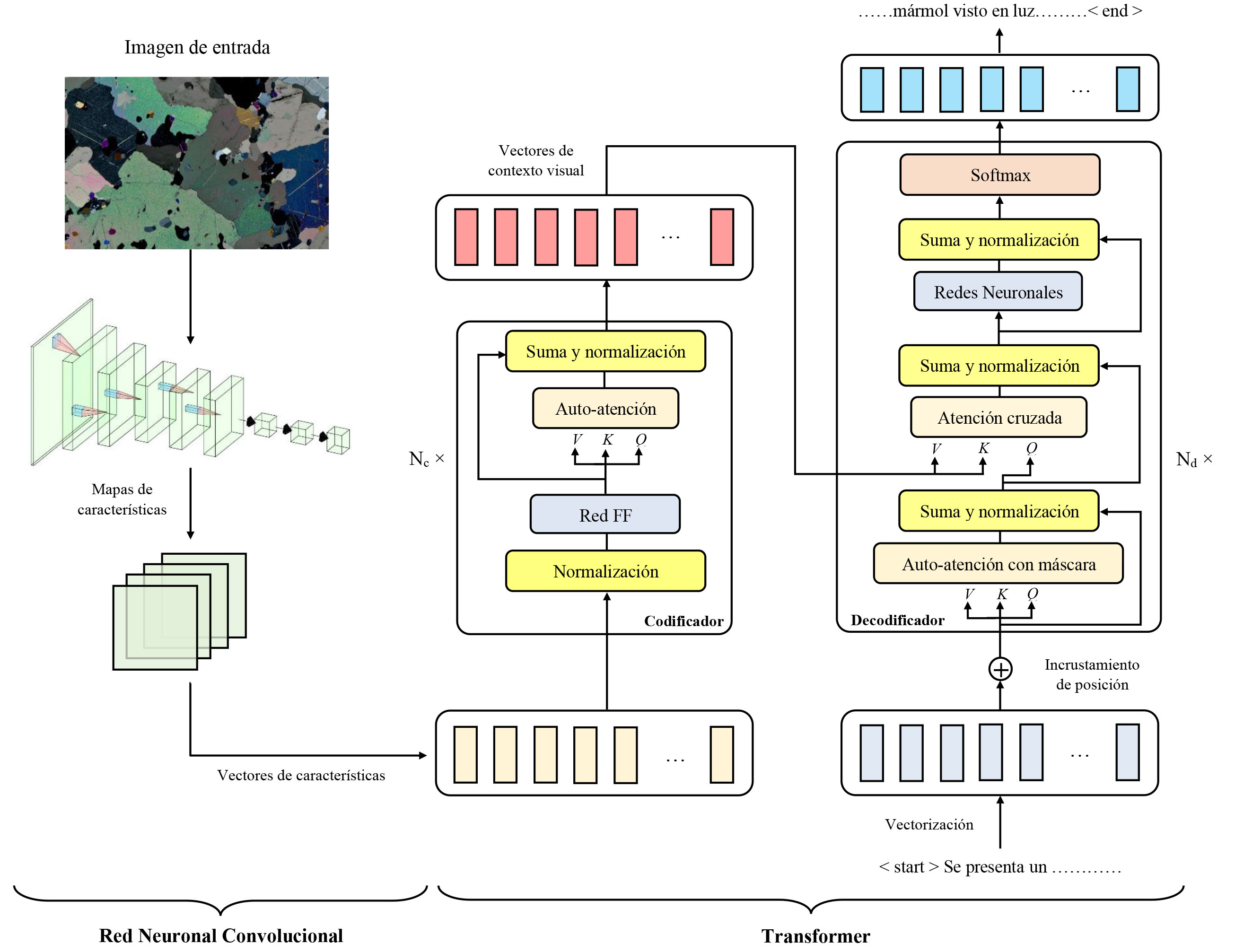}
\caption{Arquitectura del modelo de descripción textual automática.}
\label{fig:Mod_Descrp}
\end{figure}

La arquitectura del modelo y el código correspondiente han sido adaptados para trabajar con nuestro dataset de imágenes de láminas delgadas de roca y sus descripciones textuales. Esta arquitectura consta de dos partes principales: una CNN preentrenada encargada de extraer las características de las imágenes y una red transformer responsable de la generación de la descripción textual (Figura \ref{fig:Mod_Descrp}).

%Figura 4 Modelo de descripcion

En primera instancia, la imagen de entrada con una resolución de 499x499 píxeles y 3 canales de color (RGB) es procesada por un modelo preentrenado con la base de imágenes ImageNet \cite{imagenet}. Se utiliza la técnica de \textit{transfer learning} pues los filtros o pesos aprendidos son útiles para extraer los elementos visuales más importantes. Como resultado, se generan vectores de características que constituyen una versión comprimida y representativa de la imagen de entrada. Por lo señalado en la publicación de \citeA{xu2022deep} respecto al rendimiento de diferentes CNNs en diversos datasets, así como la revisión de los resultados de otros autores (\citeA{de2020petrographic}, \citeA{polat2021automatic}, \citeA{huang2023rock}), en este trabajo vamos a comparar el rendimiento de siete CNNs bien conocidas: EfficientNetB0, EfficientNetB7, EfficientNetV2L, Xception, InceptionResNetV2, ResNet50 y DenseNet121, cada una combinada con la misma red transformer. No se modificarán los hiperparámetros de las CNNs, el propósito es aprovechar modelos probados por otros autores y determinar el mejor para secciones delgadas de roca.
    
Una vez obtenidas las características de la imagen, se emplea una red transformer para asociar estas características con las descripciones textuales.
Aunque un transformer comúnmente maneja secuencias de texto de entrada y salida, en nuestro caso la entrada se trata de vectores de características de tipo visual que representan una imagen, mismas que son recibidas y preparadas por un bloque codificador para ser relacionadas con la secuencia de texto de la descripción ingresada al bloque decodificador. \citeA{Nain2021} establece la cantidad de bloques de ambas partes a $N_c = N_d = 1$. 

El bloque codificador procesa las características de la imagen de entrada y genera una representación que se utilizará como contexto para generar la descripción textual correspondiente. Específicamente, el codificador lleva a cabo los siguientes pasos:

        \begin{itemize}
            \item La normalización de la entrada (vectores de características), lo cual es común en redes neuronales para estandarizar los valores ayudando en el entrenamiento y la estabilidad del modelo.
             
            \item Una red neuronal con función de activación tipo ReLu redimensiona las entradas de tipo visual a un tamaño que posteriormente coincida con el tamaño de las entradas de tipo textual. 
            
            \item Se aplica una capa de auto-atención para establecer el grado de relación entre los diferentes elementos de la entrada. Esto se consigue por medio de la interacción de tres versiones de cada uno de los vectores de entrada denominados \textit{Query} (Q), \textit{Key} (K) y \textit{Value} (V). Puesto que se trata de determinar en qué elemento poner mayor o menor atención, es un problema de búsqueda donde la consulta Q se opera con cada clave K, obteniendo pesos de atención que ponderan todos los vectores de valor V, los cuales son combinados en uno solo. Esto se efectúa para cada vector de entrada.   
            
            \item Las conexiones residuales entre las capas ayudan a mantener información de las entradas originales y evitar el desvanecimiento del gradiente. Los valores obtenidos son normalizados antes de generar el resultado final.
            
            \item Como último paso, se obtienen vectores de contexto visual que contienen información sobre la relación de cada característica de la imagen con el resto.
            
        \end{itemize}

Por otra parte, el decodificador genera las predicciones de la descripción textual a partir de la descripción textual conocida y la salida del codificador. Para tal fin, se realizan los siguientes pasos:

        \begin{itemize}
            \item La descripción textual debe ser convertida en vectores de palabras antes de ingresar al decodificador. Estos pasan por el proceso de incrustamiento de posición  (\textit{positional embedding}) generando nuevos vectores que contienen la posición de la palabra dentro de la secuencia de la descripción textual.

            \item Puesto que el decodificador recibe y genera las palabras una por una, se aplica un tipo de auto-atención con máscara, es decir, en cada paso se consideran las palabras generadas hasta ese momento ocultando las posteriores. Dicho de otro modo, el decodificador crea el contexto de las palabras previas para predecir la próxima palabra de la frase.
%El resultado es un vector de contexto sintáctico y semántico necesario para la predicción de la siguiente palabra.
            \item En la capa de atención cruzada, el vector de contexto de la etapa anterior sirve como consulta (Q) sobre los vectores de salida del codificador, los cuales se convierten en la clave (K) y el valor (V). En este paso es donde la red aprende la relación entre las descripciones y las características de las imágenes.

            \item Las conexiones residuales y la normalización tienen el mismo propósito que en el codificador, es decir, ayudar a preservar la información de las entradas originales, mantener los valores dentro de un rango conveniente y así evitar el problema de desvanecimiento o explosión del gradiente.
            %Esto se aplica en los datos obtenidos en la incrustación de posición y en cada sub-capa de atención.            
       
            \item Dos capas densas completamente conectadas, mismas que son redes neuronales tradicionales con función de activación ReLU y que, en este caso, sirven para reproyectar el resultado anterior a las dimensiones del incrustamiento de entrada.
            
            \item Finalmente, se aplica una capa densa con la función \textit{Softmax} que convierte el resultado anterior en una distribución de probabilidad sobre el vocabulario, cuyo valor máximo corresponde a la palabra de salida.
            %, obteniendo la respectiva descripción de la imagen.
       \end{itemize}

El funcionamiento del modelo es de manera iterativa ya que el decodificador genera como salida la descripción de texto secuencialmente palabra por palabra empezando con el token \textit{<start>} y culminando con el token \textit{<end>}.
Este proceso requiere que el modelo pase por una fase de entrenamiento, en la cual se compara la palabra generada por el modelo con la palabra esperada, lo que inicia un proceso de reducción del error para ajustar los parámetros del modelo de manera óptima.
Estos pasos se llevan a cabo una vez que el modelo descrito en esta sección ha sido implementado y puesto en práctica. A continuación, se explican en detalle estos procedimientos.

\section{Experimentación y resultados}
\label{sec:experiments}
Esta sección se dedica a la parte experimental del proyecto y sus resultados. Consiste principalmente del entrenamiento del modelo descrito en la sección anterior y su respectiva evaluación. Previamente, se mencionan las especificaciones técnicas de la plataforma computacional utilizada, la preparación conveniente del dataset para el entrenamiento, así como algunos detalles de la implementación.

\subsection{Plataforma computacional}
Aprovechamos la plataforma \textit{Google Colaboratory} debido a ventajas como: libre uso, basada en la Web, permite acceder de forma remota a recursos computacionales de altas prestaciones, popular para proyectos de aprendizaje automático y con soporte para las herramientas más utilizadas en este ámbito.
En cuanto al hardware, sus características principales son: un procesador Intel\textregistered\ Xeon\textregistered\ 2.00 GHz, un disco duro con capacidad de 166.8 GB, una GPU NVIDIA Tesla V100 con RAM de 16GB y una RAM del sistema de 51 GB.
En cuanto al software, un sistema operativo Linux Ubuntu 22.04.3 LTS y para la implementación del modelo utilizamos un notebook de programación a través de un navegador y el lenguaje de programación Python. Además, las librerías de deep learning Keras y TensorFlow, y otras librerías de soporte como NumPy para el manejo de arreglos numéricos y Matplotlib para la visualización de gráficas.

\subsection{Preparación del dataset}
\label{subsec:datasetpreprocessing}

\begin{figure}[!htb]
\centering
\includegraphics[width=0.9\textwidth]{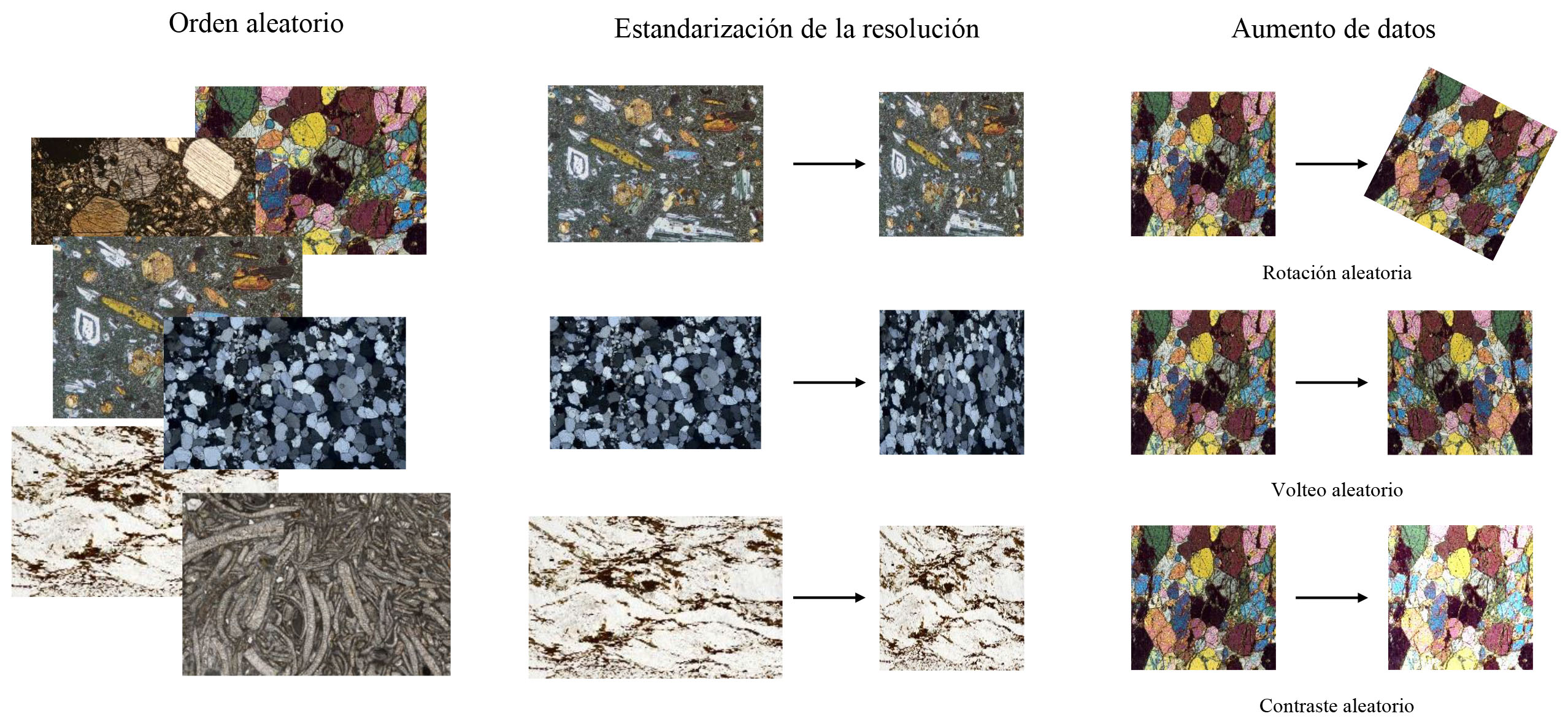}
\caption{Preparación del conjunto de imágenes para el entrenamiento.}
\label{fig:prep_IMG}
\end{figure}

Nuestro dataset es el recurso principal para el entrenamiento del modelo de descripción de rocas. El conjunto de imágenes debe ser preparado de acuerdo con el código propuesto por \citeA{Nain2021}. Debido a que las categorías están organizadas de forma secuencial, primero se asigna un orden aleatorio. Después, se estandariza la resolución de las imágenes a 499x499 píxeles. Para aumentar de manera artificial la cantidad de datos se aplican tres operaciones: rotación aleatoria, volteo aleatorio y contraste aleatorio (Figura \ref{fig:prep_IMG}).

%Figura 3 Preparacion de imágenes

Respecto a las descripciones, se debe modificar el tamaño del vocabulario o corpus de palabras, así como la longitud máxima de la secuencia, para este caso será de 300 y 60, respectivamente.
Las descripciones pasan por un proceso de vectorización en el que se requiere hacer lo siguiente: primero se debe estandarizar todas las descripciones, en donde todas tendrán letras minúsculas y se mantendrán los signos de puntuación coma ``,'' y punto ``.''. Luego se requiere tokenizar a las descripciones, es decir, separarlas palabra por palabra, cada una como un nuevo vector. Se realiza la conversión de palabra a número asignando un vector índice para cada token. Con este proceso las descripciones están listas y serán los datos de entrada del decodificador transformer.

Por último, el conjunto de datos, que contiene un total de 5600 imágenes y sus respectivas descripciones, es dividido en un 80\% para el entrenamiento y un 20\% para la validación.

\subsection{Entrenamiento}
\label{subsec:training}

Debido a que utilizamos un aprendizaje automático de tipo supervisado, el entrenamiento busca que el modelo aprenda el mapeo de las entradas con las salidas conocidas de antemano. Durante este proceso, el modelo ajusta sus pesos o parámetros de manera iterativa a partir del conjunto de datos de entrenamiento. Luego, el modelo es capaz de realizar predicciones en otros datos.

Debemos recordar que el modelo está compuesto de dos partes. La primera es una CNN que corresponde a un modelo preentrenado restringido a la extracción de características de las imágenes por lo que no requiere entrenamiento.
Por tanto, el entrenamiento únicamente se efectúa en el codificador y decodificador del transformer. Previamente, es importante la elección de los \textit{hiperparámetros}, los cuales son valores de configuración seleccionados según diversas recomendaciones y pruebas. Estos se resumen en la Tabla \ref{Tabla:Training_hyp}.

\begin{table}[!htb]
\centering
\caption{Hiperparámetros para el entrenamiento del modelo de descripción automática.}
\renewcommand{\arraystretch}{1.5}
\begin{tabular}{cccl} 
\cline{1-3}
\textbf{Hiperparámetro} & \textbf{Valor} & \textbf{Bloque} &  \\ 
\cline{1-3}
Dimensión del embedding & 512 & \multirow{2}{*}{Codificador y Decodificador} &  \\
Dimensión de la capa densa & 512 &  &  \\ 
\cline{1-3}
Número de cabezales & 1; 2 & \multirow{2}{*}{Codificador; Decodificador} &  \\
Dropout & 0.0; 0.1, 0.3, 0.5 &  &  \\ 
\cline{1-3}
Batch & 64 & \multirow{6}{*}{Modelo compilado} &  \\
Función de pérdida & Cross entropy &  &  \\
Taza de aprendizaje & 0.0001 &  &  \\
Optimizador & Adam &  &  \\
Métricas & acc \& loss &  &  \\
Épocas & 50 &  &  \\
\cline{1-3}
\end{tabular}
\label{Tabla:Training_hyp}
\end{table}

El entrenamiento se rige por los hiperparámetros.
La dimensión de 512 para el \textit{embedding} equivale al tamaño de las representaciones numéricas de las características de la imagen y las palabras.
Las capas densas realizan transformaciones lineales para lograr la dimensión requerida (512) dentro de la red.
La capa de atención del codificador tiene un único cabezal, mientras que en el decodificador se tienen dos, lo que permite capturar relaciones más complejas entre las palabras de la secuencia de entrada.
La tasa de abandono (\textit{dropout}) ayuda a regularizar el modelo desactivando aleatoriamente un porcentaje de conexiones durante el entrenamiento para prevenir el sobreajuste. Se aplican los valores de 0 y 0.1 para las capas de atención del codificador y decodificador, respectivamente; mientras que para las dos capas densas, los valores son 0.3 y 0.5, respectivamente.

La predicción del decodificador es la próxima palabra de la descripción textual y se genera mediante la auto-atención sobre las palabras previamente generadas y la atención cruzada sobre los vectores de contexto visual producidos por el codificador.
Esta predicción se compara con la respuesta conocida utilizando la función de pérdida \textit{Cross Entropy}, la cual es habitual en problemas de clasificación.
El error calculado es obtenido para grupos de 64 datos (\textit{batch}) y se reduce utilizando el optimizador \textit{Adam}, el cual es una versión mejorada del algoritmo del descenso del gradiente. Se aplica \textit{backpropagation} para la actualización de los parámetros del transformer desde las capas más cercanas a la salida hacia las capas más próximas a la entrada. Este es un proceso iterativo que ocupa 50 épocas. En la primera época se realiza un calentamiento, donde la tasa de aprendizaje parte de 0 y aumenta gradualmente hasta alcanzar el valor especificado, permitiéndole hacer mayores ajustes en las etapas iniciales y cambios más detallados cuando los parámetros están cerca de sus valores óptimos \cite{Martin_Learn_rate}.
Además, se utiliza un mecanismo llamado \textit{Early Stopping} para detener el entrenamiento una vez se haya alcanzado un valor límite en la precisión y la pérdida, esto puede ser útil para evitar un sobreajuste (\textit{overfitting)} \cite{Dagostino_Early_stop}.
El entrenamiento se detendrá si la pérdida de validación no mejora durante 3 épocas consecutivas y el modelo restaurará los mejores pesos al detenerse.

La evolución y el comportamiento del entrenamiento se pueden analizar mediante las \textit{curvas de aprendizaje}. Esta gráfica se compone de un eje horizontal para el transcurso de las épocas y un eje vertical para la precisión y pérdida del modelo (\textit{acc} y \textit{loss}). \citeA{Nain2021} define sus propias métricas a partir de la librería \textit{keras.metrics.Mean} para obtener la precisión y pérdida promedio comparando las descripciones auténticas y aquellas resultado de la predicción.

%Figura 4. Entrenamiento de todos los modelos
\begin{figure}[!htb]
\centering
\includegraphics[width=0.9\textwidth]{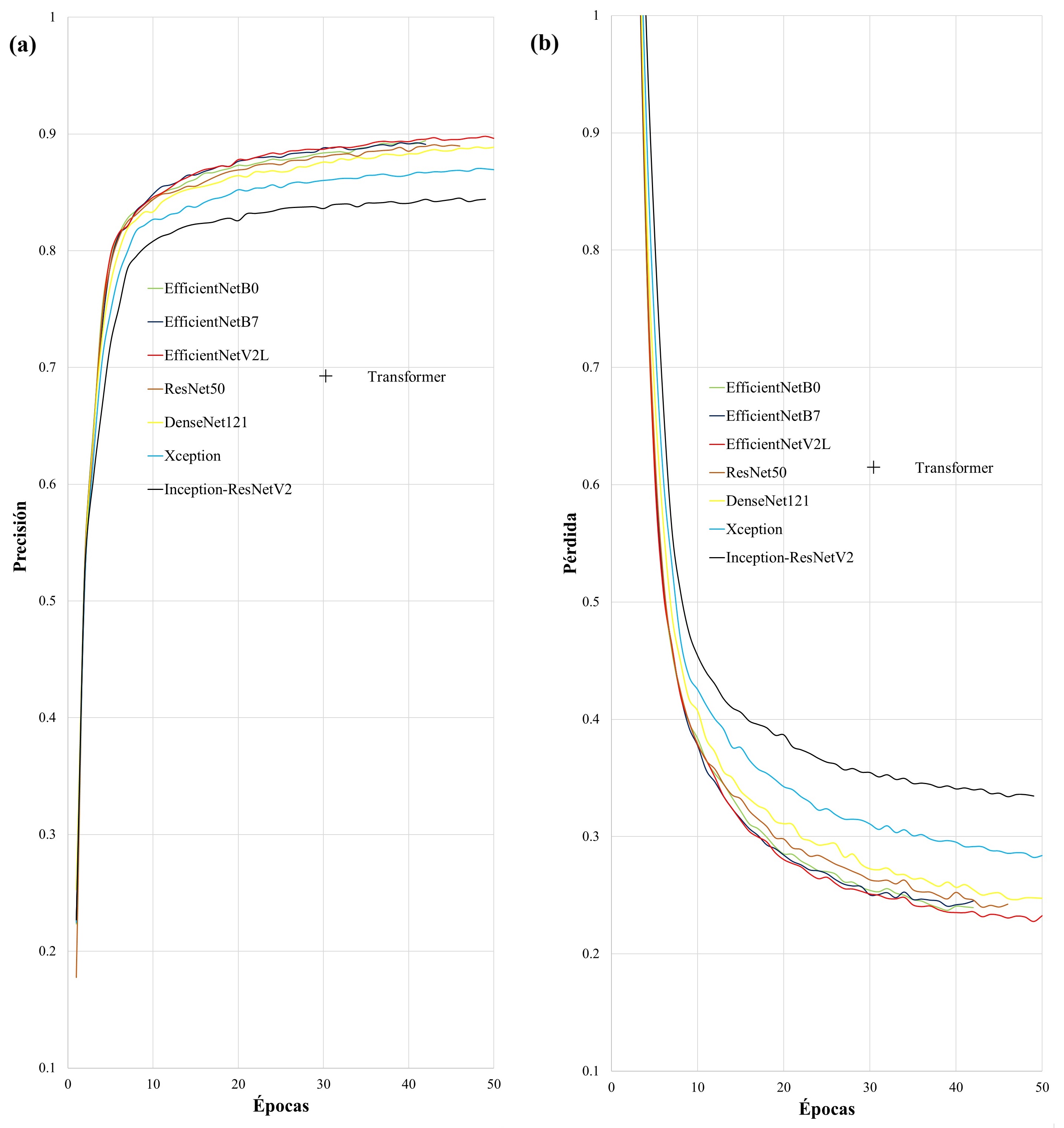}
\caption{Siete combinaciones de CNN y Transformer en el conjunto de validación. Curvas de aprendizaje de: (a) Precisión, y (b) Pérdida o error.}
\label{fig::val_los_todos}
\end{figure}

Las curvas de aprendizaje del entrenamiento en el conjunto de validación para cada una de las siete combinaciones de CNN preentrenada y transformer se presentan en la Figura \ref{fig::val_los_todos}, tanto para la métrica de precisión (a) como para la pérdida (b). Se observa que en un inicio todos los modelos tienen un comportamiento similar. A partir de la tercera época, los modelos con DenseNet121, Xception e Inception-ResNetV2 se separan de la tendencia. De estos tres, DenseNet121 y Xception completaron las 50 épocas por lo que es posible que al aumentar este hiperparámetro mejoren su rendimiento. Luego de la décima época, se marcan diferencias con las otras: EfficientNetV2L, EfficientNetB7, EfficientNetB0 y ResNet50, de las cuales, solamente EfficientNetV2L completó el entrenamiento. Por tanto, los tres mejores resultados de precisión fueron obtenidos al combinar EfficientNetV2L, EfficientNetB0 y EfficientNetB7 con la red transformer. Los valores máximos son de 0.898, 0.894 y 0.892, respectivamente. Estos valores superan por más de la mitad al obtenido por \citeA{Nain2021}, al aplicar este modelo en el conjunto de datos de \textit{Flickr8k}.

\begin{figure}[!htb]
\centering
\includegraphics[width=0.9\textwidth]{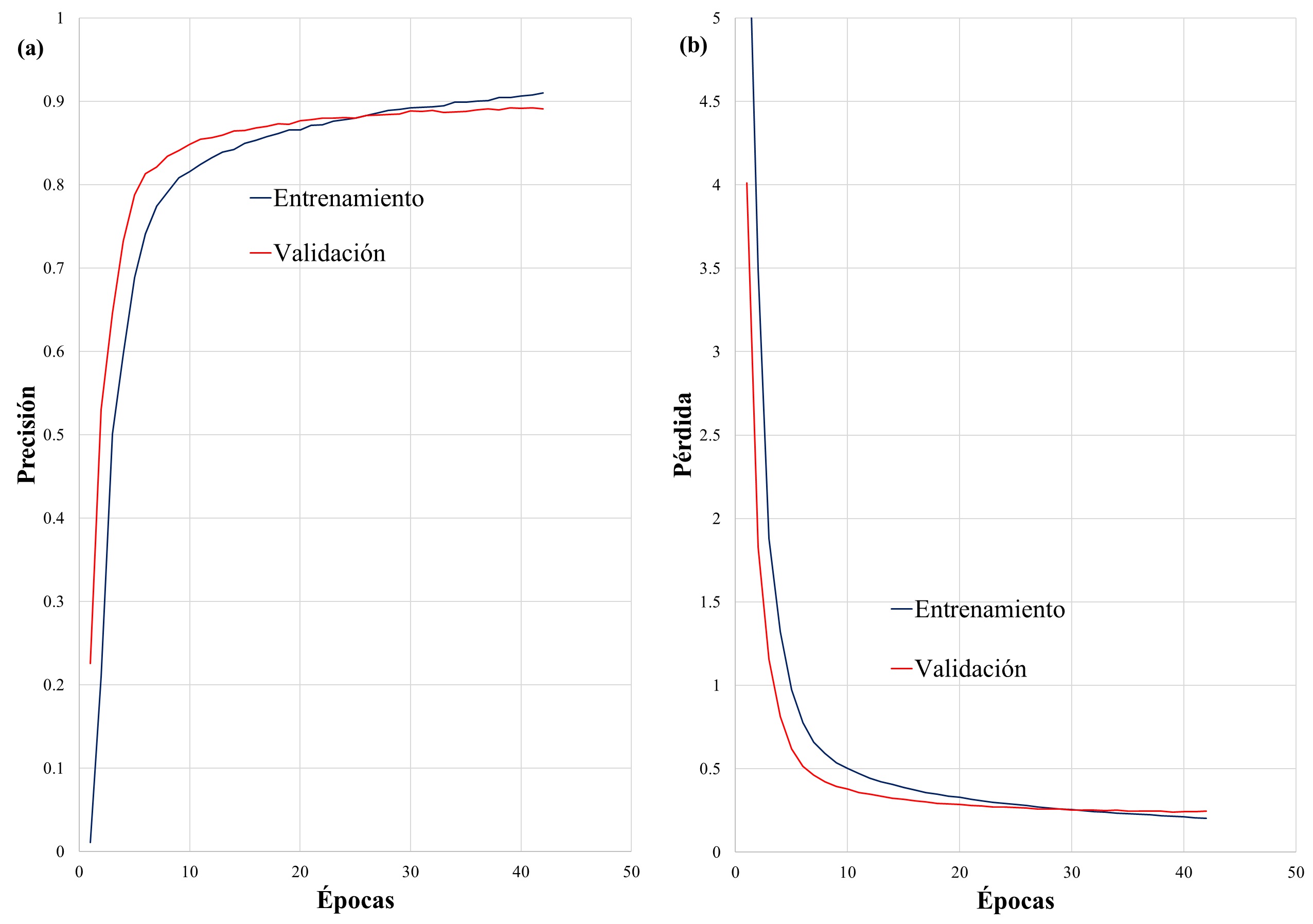}
\caption{Curvas de aprendizaje de precisión (a) y pérdida (b) para el conjunto de validación del modelo de descripción que utiliza EfficientNetB7.}
\label{fig:val_los_b7}
\end{figure}

Las curvas de aprendizaje de los modelos entrenados tienen un comportamiento similar. Para explicarlo, se toma como ejemplo la combinación de EfficienNetB7 y el transformer (Figura \ref{fig:val_los_b7}). Se debe notar que hasta la época 27 los valores de precisión del conjunto de validación son mayores al conjunto de entrenamiento y menores hasta la época 31 en los valores de pérdida. El entrenamiento del conjunto original de datos de \citeA{Nain2021} también presenta aquel comportamiento. Esto se puede atribuir a que el conjunto de validación no tiene la misma complejidad que el de entrenamiento. Para intentar solucionarlo se cambió la partición de los datos, considerando un 70\% para el conjunto de entrenamiento y 30\% para el de validación. El entrenamiento entregó resultados similares. \citeA{Frederik_drop} indica que el hiperparámetro de dropout influye en las curvas de aprendizaje, al añadir este valor las curvas son menos regulares ya que no se utilizan todos los datos. Al eliminar estos valores del modelo, en la octava época la precisión del conjunto de entrenamiento superó al de validación. Además, el entrenamiento terminó en la época 32. 

%Figura 5. Entrenamiento del mejor modelo

\subsection{Evaluación BLEU de las descripciones}
Los valores de precisión y pérdida son buenos indicadores acerca del rendimiento del modelo. Sin embargo, se requiere complementar la evaluación utilizando la métrica BLEU, tanto para la descripción textual total como para cada una de las partes que la componen.
La métrica BLEU es utilizada para evaluar sistemas de Procesamiento del Lenguaje Natural (PLN), en particular, sistemas de traducción automática y generación de lenguaje natural. El resultado de está métrica está en el rango de 0 a 1, lo que permite asignar un valor cualitativo según la escala de \citeA{reiter2018BLEU} (Tabla \ref{Tabla:BLEUescala}). 

\begin{table}[!htb]
\centering
\caption{Escala para la valoración cualitativa de la métrica BLEU.}
\renewcommand{\arraystretch}{1.3}
\begin{tabular}{cc} 
\hline
\textbf{Valor numérico} & \textbf{Valor cualitativo} \\ 
\hline
0.85-1.00 & Alto \\
0.70-0.85 & Medio \\
0.00-0.70 & Bajo \\
\hline
\end{tabular}
 \label{Tabla:BLEUescala}
\end{table}

Se escogieron las tres combinaciones de CNN y transformer con mayor precisión en el conjunto de validación y que superen el 85\%. Estos modelos tienen a las siguientes CNNs: EfficientNetV2L, EfficientNetB0 y EfficientNetB7.
Para la aplicación de la métrica BLEU, utilizamos un conjunto de prueba externo con 56 imágenes, es decir, 4 imágenes de cada categoría. Se evaluaron las descripciones generadas por los tres modelos, en total 168 descripciones. La Tabla \ref{Tabla:BLEUB7B0V2L} presenta los resultados de la métrica BLEU tomando en cuenta el tipo de roca y el modelo combinado con su respectiva CNN. Cada tipo de roca tiene 4 descripciones, el valor indicado representa un promedio.

\begin{table}[!htb]
\centering
\caption{Resultados de BLEU para el modelo de descripción automática usando EfficientNetB0, EfficientNetB7 y EfficientNetV2L.}

\renewcommand{\arraystretch}{1.5}
\begin{tabular}{cccc} 
\hline
 & \textbf{EfficientNetB0} & \textbf{EfficientNetB7} & \textbf{EfficientNetV2L} \\ 
\hline
\textbf{Riolita} & 0.791 & 0.819 & 0.779 \\
\textbf{Andesita} & 0.792 & 0.749 & 0.768 \\
\textbf{Basalto} & 0.436 & 0.551 & 0.407 \\
\textbf{Granito} & 0.594 & 0.626 & 0.603 \\
\textbf{Diorita} & 0.742 & 0.753 & 0.682 \\
\textbf{Gabro} & 0.923 & 0.925 & 0.881 \\
\textbf{Ultramáfica} & 0.839 & 0.861 & 0.717 \\
\textbf{Filita} & 0.622 & 0.586 & 0.363 \\
\textbf{Esquisto} & 0.620 & 0.626 & 0.581 \\
\textbf{Gneis} & 0.791 & 0.794 & 0.761 \\
\textbf{Mármol} & 0.878 & 0.822 & 0.849 \\
\textbf{Arenisca} & 0.541 & 0.610 & 0.701 \\
\textbf{Lutita} & 0.457 & 0.618 & 0.333 \\
\textbf{Caliza} & 0.465 & 0.608 & 0.646 \\ 
\hline
\textbf{Promedio} & 0.678 & 0.710 & 0.648 \\
\hline
\end{tabular}
 \label{Tabla:BLEUB7B0V2L}
\end{table}

Se observa que el mayor valor BLEU lo obtiene el modelo con EfficientNetB7 (0.71), lo que en la escala cualitativa de \citeA{reiter2018BLEU} es de nivel medio. En este modelo, los resultados más altos son para la categoría de Gabro y Ultramáfica, resultados medios para las categorías Riolita, Andesita, Diorita, Gneis y Mármol, y bajos para Basalto, Granito, Esquisto, Filita, Arenisca, Caliza y Lutita. Es decir, una mitad del conjunto de prueba está en un rango alto-medio y la otra mitad en bajo.

Para explicar con más detalle las capacidades y limitaciones de este modelo, se ha aplicado la métrica BLEU a cada uno de los cinco grupos de parámetros de descripción, es decir, roca y tipo de luz, textura, minerales o componentes, forma y hábito, y relieve o colores de interferencia. Estos resultados se presentan en la Tabla \ref{Tabla:BLEUB7}.

%Tabla de resultados de BLEU de cada parámetro.
\begin{table} [!htb]
\centering
\caption{Resultados de BLEU para el modelo de descripción automática usando EfficientNetB7.}

\renewcommand{\arraystretch}{1.5}
\begin{tabular}{cccccc} 
\hline
 & \begin{tabular}[c]{@{}c@{}}\textbf{Roca y}\\\textbf{tipo de luz}\end{tabular} & \textbf{Textura} & \begin{tabular}[c]{@{}c@{}}\textbf{Minerales o}\\\textbf{Componentes}\end{tabular} & \begin{tabular}[c]{@{}c@{}}\textbf{Forma~y}\\\textbf{Hábito}\end{tabular} & \begin{tabular}[c]{@{}c@{}}\textbf{Relieve o}\\\textbf{Colores de interferencia}\end{tabular} \\ 
\hline
\textbf{Riolita} & 1.000 & 0.893 & 1.000 & 0.595 & 1.000 \\
\textbf{Andesita} & 1.000 & 1.000 & 0.870 & 0.446 & 0.908 \\
\textbf{Basalto} & 0.641 & 0.893 & 0.391 & 0.327 & 0.908 \\
\textbf{Granito} & 1.000 & 0.834 & 0.318 & 0.786 & 0.832 \\
\textbf{Diorita} & 1.000 & 0.784 & 0.507 & 0.664 & 0.908 \\
\textbf{Gabro} & 1.000 & 1.000 & 1.000 & 0.893 & 0.834 \\
\textbf{Ultramáfica} & 1.000 & 0.752 & 0.904 & 0.875 & 1.000 \\
\textbf{Filita} & 0.641 & 0.938 & 0.509 & 0.520 & 0.605 \\
\textbf{Esquisto} & 0.641 & 0.750 & 0.370 & 0.560 & 0.737 \\
\textbf{Gneis} & 0.821 & 0.780 & 0.626 & 0.817 & 0.917 \\
\textbf{Mármol} & 1.000 & 1.000 & 0.821 & 0.676 & 1.000 \\
\textbf{Arenisca} & 0.821 & 0.838 & 0.522 & 0.485 & 0.908 \\
\textbf{Lutita} & 0.821 & 0.917 & 0.357 & 0.000 & 0.908 \\
\textbf{Caliza} & 0.932 & 0.784 & 0.750 & - & 0.326 \\ 
\hline
\textbf{Promedio} & 0.880 & 0.869 & 0.639 & 0.588 & 0.842 \\
\hline
\end{tabular}
 \label{Tabla:BLEUB7}
\end{table}

De manera general, los valores altos corresponden a roca y tipo de luz, y textura; los valores medios a relieve o colores de interferencia; y, los valores bajos a minerales o componentes principales, así como forma y hábito. En particular, el tipo de roca y el tipo de luz son reconocidos en la mayoría de casos. El valor más bajo se obtiene en basaltos que el modelo confunde por riolitas en dos ocasiones y en esquistos que son confundidos por gneises también en dos ocasiones.
Existen diferencias marcadas entre basaltos y riolitas, en particular respecto a su mineralogía y textura. Esta confusión se presenta en luz polarizada plana, posiblemente confundiendo las formas de los minerales por una textura vacuolar presente en ambas categorías. Respecto al esquisto y gneis, la diferencia está en la textura y el grado de metamorfismo, presentando este último cristales de mayor tamaño. Aquel parámetro no es considerado en esta descripción por lo que se esperaba aquel resultado; la confusión se presenta en una ocasión en PPL y otra en XPL. Una imagen de filita es confundida por un gneis (XPL) y en otra por un mármol (PPL). En XPL una imagen de lutita es descrita como filita, también se esperaba esta confunsión pues en ciertas ocasiones existen semejanzas en el tamaño de grano y las laminaciones de la lutita pueden ser similares a la foliación de la filita. En cuanto a las areniscas, hubo una confusión por diorita lo cual puede ser justificado hasta cierto punto por la presencia de cuarzo, feldespato potásico y biotita, cuya textura en PPL se asemeja, en aquel caso, a una diorita.
Respecto a las calizas, el modelo identifica de manera correcta a la roca, sin embargo, falla en dos ocasiones al identificar el tipo de luz. 

El parámetro de textura tiene una mayor complejidad pues son múltiples texturas que existen dentro de un mismo tipo de roca y el primer problema presente es una representación no equitativa en el dataset. A pesar de esto, se ha obtenido buenos resultados, con valores en un rango medio-alto.

El parámetro de relieve o colores de interferencia tiene también valores en el rango medio-alto, con excepción de la categoría de caliza, que como resultado de la confusión en identificar el tipo de luz describe de forma errónea el relieve y los colores de interferencia. No obstante, se puede rescatar algo positivo de esto. El parámetro de tipo de luz al inicio de la descripción está relacionado con los parámetros relieve o color de interferencia ubicados al final de la descripción, el mecanismo de atención debe dar un mayor peso a esta relación, las descripciones en PPL coinciden con relieve y aquellas en XPL coinciden con colores de interferencia. El mecanismo de atención es un avance respecto a las redes neuronales recurrentes (RNN) con memoria a largo y corto plazo (LSTM). \citeA{menendez2022reconocimiento} usan la anterior arquitectura, sin embargo, no discuten las relaciones contextuales entre sus parámetros de descripción.

Los valores más bajos corresponden a minerales o componentes principales (0.639), así como forma y hábito (0.588). Estos son los parámetros más complejos de describir. En general se puede decir que el primero tiene una relación contextual con el tipo de roca. Por ejemplo, en una riolita están presentes cristales de cuarzo y feldespato potásico. Esta asunción se observa también en aquellas imágenes de basaltos que fueron confundidas por riolitas. Es decir, el modelo aprende la mineralogía en relación al tipo de roca. Para identificar rocas ígneas se requiere obtener la composición mineral modal, en rocas metamórficas minerales índice y textura, en rocas sedimentarias detríticas el tamaño de grano-composición modal y, en rocas carbonatadas se debe identificar componentes aloquímicos y ortoquímicos. Dicho de otro modo, cada tipo de roca presenta su propia particularidad, lo que limita a un modelo de descripción automática a obtener características detalladas y precisas de éstas. En este caso, el modelo está memorizando esta asociación.

\section{Discusión de resultados}
\label{sec:discussion}
Durante la selección de la CNN se ha notado cierta incertidumbre respecto a la cantidad de capas de las mismas. Es decir, CNNs con pocas capas tendrían un flujo de propagación hacia adelante y hacia atrás más eficiente, pero las características extraídas no serían representativas. Por otro lado, CNNs con más capas tienen modelos más complejos que ayudan a la extracción de características, sin embargo, serían más complicadas de entrenar en términos de costo computacional \cite{de2020petrographic}. Además, la experimentación ha demostrado que aumentar capas a una CNN no implica una mejora en la precisión.

Los trabajos de \citeA{de2020petrographic} y \citeA{polat2021automatic} coinciden en que ResNet50 resuelve el problema del desvanecimiento del gradiente, lo cual ocurre conforme se aumentan las capas de una CNN. Ambos autores obtienen los mejores resultados con aquella CNN. 
\citeA{polat2021automatic} y \citeA{xu2022deep} discuten las ventajas de usar DenseNet121. Señalan que el diseño de bloques densos con menos parámetros hace que la transmisión de características y gradientes sea más eficiente. \citeA{xu2022deep} hacen uso de las CNNs Xception e InceptionResNetV2, la primera obtiene el mejor resultado asociándolo a la convolución de profundidad separada (\textit{depth separable convolution}), lo cual hace que la CNN tenga más capas con la misma cantidad de parámetros; en cuanto a InceptionResNetV2, su segundo mejor resultado, lo asocia a la profundidad y amplitud de esta CNN.

\citeA{Nain2021} y \citeA{huang2023rock} utilizan la arquitectura EfficientNetB0 y EfficientNetB7, respectivamente. \citeA{huang2023rock} indican que el proceso de escalamiento del modelo implica cambios en la profundidad, amplitud de la arquitectura de la CNN y cambios en la resolución de la imagen; lo que ha tenido un notable impacto en mejorar la precisión. Para comparar los resultados, también se ha seleccionado un modelo más reciente, EfficientNetV2L, cuyo rendimiento con la base de datos de validación de ImageNet, se encuentra en tercer puesto considerando una precisión top-1 con 85.7\% y, en primer puesto considerando una precisión top-5 con 97.5\%. Sin embargo, se observó que el modelo con esta CNN obtiene el resultado más bajo en la evaluación BLEU.

Respecto a las descripciones textuales, una vez determinado el mejor modelo se hizo la evaluación independiente de cada una de las partes que componen la descripción. Se obtiene un resultado alto en tipo de roca y luz (0.880) y textura (0.869), un valor medio en relieve o colores de interferencia (0.842) y valores bajos en minerales o componentes principales (0.639) y, forma y hábito (0.588). El conjunto de datos está equilibrado respecto a la cantidad de imágenes de cada tipo de roca y luz, esto explicaría los buenos resultados de aquellos parámetros. Sin embargo, no es el caso del parámetro textura. A pesar de que cada tipo de roca presenta múltiples texturas, se obtiene un valor alto, lo que indica que el modelo es bueno extrayendo características para relacionarlas con este parámetro. En cuanto a los valores bajos de minerales o componentes principales y, forma y hábito, los cuales son más complejos de describir, se puede considerar que el modelo memoriza la descripción asociando el tipo de roca con los minerales y estos últimos con formas y hábitos. Para una mejor identificación de estos parámetros se requiere de otras estrategias de visión por computadora como la clasificación semántica multiclase y la detección de objetos.

\section{Aplicación Web}
\label{sec:appweb}
% Justificación
A menudo, los proyectos de aprendizaje automático se limitan a la etapa de desarrollo, ya que desplegar los modelos obtenidos en un entorno de producción puede ser una tarea técnica compleja y costosa en tiempo y recursos. En cambio, nuestro principal objetivo es ofrecer una herramienta interactiva que resulte beneficiosa tanto para el ámbito académico como profesional, así como para usuarios en general.
Por tanto, nos hemos propuesto hacer disponible a nuestro modelo ya entrenado para cualquier persona con interés en el estudio de rocas y en cualquier parte del mundo para generar descripciones textuales y verbales a partir de imágenes suministradas por ellos mismos.

% Análisis
Son varios los requerimientos que debe satisfacer el producto final como la facilidad de uso, accesibilidad, apariencia atractiva y contenido informativo basado en texto, imágenes y animaciones que deje claro los alcances y limitaciones del modelo. Tales requisitos se ajustan a una intefaz sencilla, dinámica y universal, por lo que hemos considerado desplegar nuestro modelo mediante la implementación de una aplicación Web.

% Implementación e infraestructura
Esta implementación consta de dos partes: el despliegue de la aplicación y la creación y diseño del sitio Web. Las plataformas utilizadas para el desarrollo y puesta en práctica de nuestra aplicación Web son todas libres y de código abierto (open source). La arquitectura de la solución es de tipo cliente/servidor. De manera general, se requiere un servidor Web que aloje y cargue el modelo para atender peticiones de los usuarios realizadas a través de una dirección URL especificada en el navegador de su computadora o dispositivo móvil.

\subsection{Despliegue de la aplicación}
%Servidor de la aplicación
La elección del servidor para la aplicación es crucial pues tiene impacto en su rendimiento y disponibilidad.
Existen diversas plataformas que pueden ser alternativas adecuadas para poner a nuestro modelo a trabajar en la nube como Heroku, Anvil, Streamlit, entre otras.
La más adecuada debe proporcionar herramientas intuitivas para la configuración y la implementación de la aplicación, debe ser capaz de gestionar eficientemente los picos de tráfico de usuarios facilitando así un uso eficiente. De igual importancia es la seguridad, con mecanismos para proteger tanto la aplicación como los datos de los usuarios.

En este caso se escogió la plataforma de Streamlit, la cual no requiere conocimientos avanzados de programación, lo que permite a los desarrolladores crear y desplegar aplicaciones de IA de manera rápida y eficiente. La aplicación se crea en un simple paso en donde se coloca la URL del repositorio de GitHub que contiene nuestro modelo y un archivo Python (\textit{.py}) donde se configura los elementos de la aplicación, por ejemplo, la opción para usuarios externos de cargar archivos.

Para configurar y administrar el repositorio de GitHub es necesario utilizar la versión local para computadoras. Este repositorio contiene el conjunto de datos de imágenes y descripciones, el código del modelo de descripción automática, la aplicación en Python, así como los requisitos de la aplicación en un archivo de texto (\textit{.txt}). El modelo de descripción automática es dividido en dos archivos: uno para su arquitectura y otro para los pesos obtenidos durante la fase de entrenamiento. Los pesos fueron guardados con la extensión H5. Respecto al archivo de requisitos, éste incluye las librerías usadas. 
% Voz
En el archivo de la aplicación se debe realizar la respectiva configuración para convertir la descripción textual a voz. Esto es posible mediante el uso de \textit{Google Text-to-Speech} (GTTS), la cual es una API que tiene la ventaja de ser de acceso libre, sin límite de conversión de caracteres. Esta descripción verbal puede ser guardada con la extensión MP3. Tanto el audio como la descripción textual son archivos temporales y son eliminados una vez se deje de usar la aplicación.

\subsection{Diseño del sitio Web}
%Sitio web
Se escogió la popular plataforma de desarrollo Wix\footnote{\url{www.wix.com}}, la cual permite de manera simple y asisitida a los usuarios crear y administrar sus propios sitios Web. En esta plataforma se ha diseñado un sitio compuesto de varias páginas para que los usuarios se registren, accedan a la información más relevante sobre el modelo, tengan acceso al conjunto de datos de imágenes y descripciones, e interactúen con la aplicación, tal como se muestra en la Figura \ref{fig:arq}.
Estas funcionalidades se detallan a continuación:

\begin{figure}[!htb]
    \centering
    \includegraphics[width=0.9\textwidth]{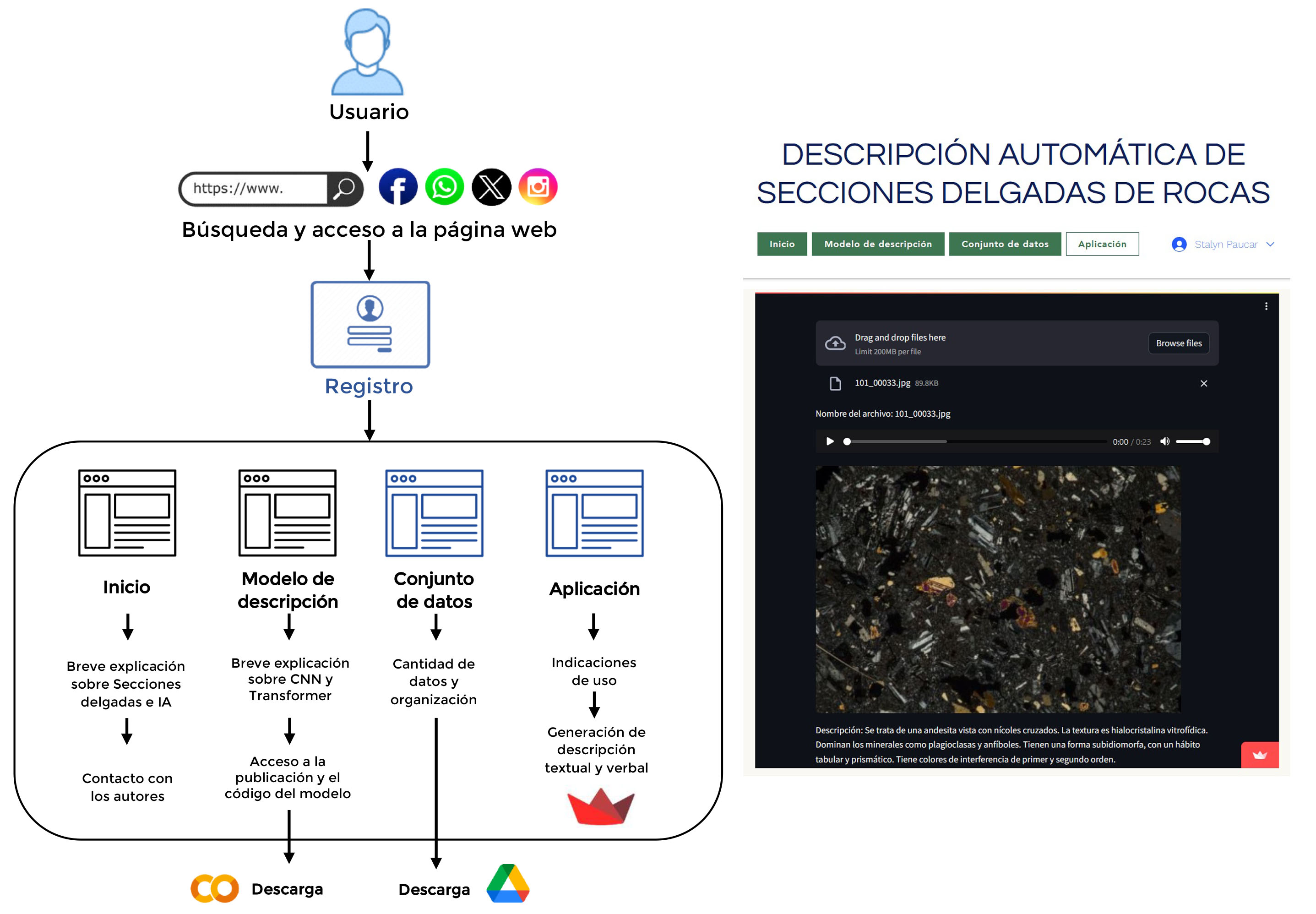}
    \caption{Funcionalidades del sitio Web (izquierda) y una captura de pantalla de la aplicación para la descripción textual y verbal de láminas de roca (derecha).}
    \label{fig:arq}
\end{figure}
\begin{enumerate}
    \item Acceso al sitio Web: los usuarios pueden acceder al sitio de diferentes maneras, ya sea mediante un buscador, enlace directo o a través de diferentes redes sociales.    
    \item Registro de usuarios: los visitantes del sitio pueden escoger la opción para crear una cuenta de acceso, lo que constituye un primer filtro para controlar el uso de la aplicación. Los datos solicitados no comprometen al usuario y respetan su privacidad. Se realiza el registro mediante una cuenta de correo y datos adicionales como nombre, apellido, país de origen, institución y edad. El registro es obligatorio para acceder al conjunto de datos y la aplicación.    
    \item Inicio: esta es una página de bienvenida. Se explica de forma breve la importancia de las secciones delgadas de rocas y el auge de la IA para analizar múltiples datos usados en Ciencias de la Tierra. También se presenta una sección para ponerse en contacto con los autores.    
    \item Modelo de descripción: esta página presenta un resumen del trabajo desarrollado y la metodología utilizada, indicando en qué consiste el modelo, sus capacidades y limitaciones. El usuario tiene acceso a esta publicación y también puede descargar el código en la plataforma de Google Colaboratory.
    \item Conjunto de datos: aquí se explica cómo se han organizado las imágenes de secciones delgadas de rocas y sus respectivas descripciones. Se encuentra disponible el enlace para la descarga del dataset. El usuario debe explicar el uso que dará al conjunto de datos y posteriormente accederá a un enlace de Google Drive.
    \item Aplicación: Wix permite la inserción de la aplicación generada en Streamlit mediante un enlace. Los usuarios pueden cargar sus propias imágenes de muestras de roca, visualizar y escuchar la respectiva descripción. Además, se añade un cuadro de comentarios para que el usuario evalúe la descripción.
\end{enumerate} 
 Estas funcionalidades pueden ser aprovechadas desde una computadora convencional, así como dispositivos móviles.
% Pruebas
El producto final se encuentra disponible al público en general mediante la dirección URL: \url{stalynpaucar271828.wixsite.com/auto-descripcion}. En la Figura \ref{fig:arq} también se observa una prueba de la descripción automática a partir de una imagen de muestra de roca cargada por el usuario.

%###################################################################
%-----------------------CONCLUSIONES---------------------------
%###################################################################
\section{Conclusiones y trabajo futuro}
\label{sec:conclusion}

Hemos presentado el desarrollo de un sistema de descripción automática de tipo textual y verbal de láminas delgadas de roca utilizando herramientas de inteligencia artificial.
El uso de estas arquitecturas de redes neuronales modernas es fundamental para implementar un sistema automatizado para la descripción de estas imágenes.
La solución propuesta está basada en la combinación de una red neuronal convolucional y una red transformer para el reconocimiento de las características de la imagen y su descripción, respectivamente. De tal forma, se integran dos importantes campos como la visión por computadora y el procesamiento del lenguaje natural. Además, se ha añadido la capacidad de convertir texto a voz con el fin de ampliar la funcionalidad, permitiendo no sólo un resultado visual, sino también verbal.

Uno de nuestros principales aportes es la creación de un dataset compuesto de 5600 imágenes de secciones delgadas organizadas en 14 tipos de rocas y una base de descripciones textuales, lo cual ha sido el recurso principal para el aprendizaje supervisado del modelo de descripción automática. Este recurso también puede ser de utilidad para otros trabajos relacionados.

La experimentación consideró siete CNNs preentrenadas del estado del arte: EfficientNetB0, EfficientNetB7, EfficientNetV2L, Xception, InceptionResNetV2, ResNet50 y DenseNet121, cada una combinada con la red transformer.
Se determinó que las diversas versiones de EfficientNet presentan los mejores resultados de precisión y pérdida.
Para escoger la mejor opción de CNN, se ha realizado la evaluación en un conjunto de prueba externo de 56 imágenes aplicando la métrica BLEU. El resultado fue que el modelo que combina EfficientNetB7 y la red transformer tiene el mejor valor con un 0.71 de precisión, esto sería considerado como un resultado de calidad aceptable. 

El modelo obtenido se ha desplegado a través de un sitio Web propio que está destinado al uso del público en general e incluye funciones para el registro y control de los usuarios, la comunicación con los autores para solventar dudas e inconvenientes al momento de usar este producto, la carga de imágenes y el despliegue de la descripción textual y verbal, y la descarga del dataset. Las tecnologías de implementación de esta aplicación Web incluyen las plataformas GitHub, Streamlit y WIX. Todas ellas tienen la ventaja de ser de acceso libre y permiten que la aplicación se encuentre en línea de forma continua. 

Existe potencial para mejorar el modelo al incluir una mayor cantidad de imágenes que exhiban un equilibrio en los parámetros de descripción. Además, es importante destacar que el análisis se limitó a rocas frescas, lo que sugiere la necesidad de futuras investigaciones que consideren la inclusión de minerales de alteración. Por otro lado, este estudio se centró exclusivamente en el uso de luz transmitida para el análisis petrográfico, lo que deja abierta la oportunidad para investigaciones adicionales que exploren el uso de luz reflejada. Este enfoque podría ser especialmente relevante para estudiar minerales de mena. También se debe trabajar en otras arquitecturas como segmentación semántica multiclase y detección de objetos. En conjunto, estas consideraciones resaltan la importancia de una investigación continua en el campo del análisis geológico y petrográfico, con el potencial de mejorar la precisión y la aplicabilidad de los modelos de inteligencia artificial en este dominio.

%###################################################################
%-----------------------REFERENCIAS-------------------------------------
%###################################################################

\newpage

\bibliographystyle{apacite}

\bibliography{Doc_Principal_APA}

\end{document}